\documentclass{article}

\usepackage{arxiv}
\usepackage{amsmath}
\usepackage[utf8]{inputenc} 
\usepackage[T1]{fontenc}    
\usepackage{hyperref}       
\usepackage{url}            
\usepackage{booktabs}       
\usepackage{amsfonts}       
\usepackage{nicefrac}       
\usepackage{microtype}      
\usepackage{lipsum}		
\usepackage{graphicx}
\usepackage{natbib}
\usepackage{doi}
\usepackage{float}
\usepackage{titlesec}

\AtBeginDocument{
  \setlength{\headheight}{22.37994pt}
  \addtolength{\topmargin}{-8.37994pt}
}

\title{Silhouette Loss: Differentiable Global Structure Learning for Deep Representations}

\date{} 					

\author{
	\href{https://orcid.org/0000-0001-7568-8784}{\includegraphics[scale=0.06]{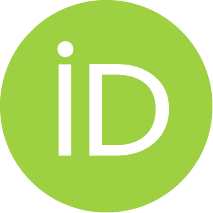}\hspace{1mm}Matheus V. Todescato} \\
	Institute of Informatics\\
	UFRGS\\
    Porto Alegre, Brazil \\
	\texttt{mvtodescato@inf.ufrgs.br} \\
    \And
	\href{https://orcid.org/0000-0002-4499-3601}{\includegraphics[scale=0.06]{orcid.pdf}\hspace{1mm}Joel L. Carbonera} \\
	Institute of Informatics\\
	UFRGS\\
	Porto Alegre, Brazil \\
	\texttt{jlcarbonera@inf.ufrgs.br} \\
}

\renewcommand{\headeright}{}

\renewcommand{\undertitle}{}
\renewcommand{\shorttitle}{Silhouette Loss: Differentiable Global Structure Learning for Deep Representations}

\hypersetup{
pdftitle={Silhouette Loss: Differentiable Global Structure Learning for Deep Representations},
pdfsubject={cs.CV},
pdfauthor={Matheus V. Todescato, Joel L. Carbonera},
pdfkeywords={Machine Learning, Convolutional Neural Networks, Image Classification},
}

\begin{document}
\maketitle

\begin{abstract}
Learning discriminative representations is a central goal of supervised deep learning. While cross-entropy (CE) remains the dominant objective for classification, it does not explicitly enforce desirable geometric properties in the embedding space, such as intra-class compactness and inter-class separation. Existing metric learning approaches, including supervised contrastive learning (SupCon) and proxy-based methods, address this limitation by operating on pairwise or proxy-based relationships, but often increase computational cost and complexity. In this work, we introduce Soft Silhouette Loss, a novel differentiable objective inspired by the classical silhouette coefficient from clustering analysis. Unlike pairwise objectives, our formulation evaluates each sample against all classes in the batch, providing a batch-level notion of global structure. The proposed loss directly encourages samples to be closer to their own class than to competing classes, while remaining lightweight. Soft Silhouette Loss can be seamlessly combined with cross-entropy, and is also complementary to supervised contrastive learning. We propose a hybrid objective that integrates them, jointly optimizing local pairwise consistency and global cluster structure. Extensive experiments on seven diverse datasets demonstrate that: (i) augmenting CE with Soft Silhouette Loss consistently improves over CE and other metric learning baselines; (ii) the hybrid formulation outperforms SupCon alone; and (iii) the combined method achieves the best performance, improving average top-1 accuracy from 36.71\% (CE) and 37.85\% (SupCon2) to 39.08\%, while incurring substantially lower computational overhead. These results suggest that classical clustering principles can be reinterpreted as differentiable objectives for deep learning, enabling efficient optimization of both local and global structure in representation spaces.
\end{abstract}

\section{Introduction}

Deep neural networks trained with the cross-entropy loss have achieved remarkable success in several visual classification tasks.
Despite its effectiveness for optimizing predictive accuracy, cross-entropy does not explicitly enforce desirable geometric properties in the learned representation space \citep{elsayed2018large,sukhbaatar2014training}.
In particular, it does not directly encourage samples from the same class to form compact clusters, nor does it ensure strong separation between classes.
As a result, learned embeddings may exhibit suboptimal structure for downstream tasks such as retrieval, transfer learning, or open-set recognition \citep{liu2016large}.

Several approaches have been proposed to address this limitation by introducing additional objectives that shape the geometry of the embedding space.
Metric learning methods \citep{moutafis2016overview} encourage semantically similar samples to lie close together while pushing dissimilar samples apart.
Prototype-based losses, such as Center Loss \citep{wen2019comprehensive} or angular-margin objectives \citep{lee2022angular}, explicitly enforce intra-class compactness and inter-class separation using class representatives.
More recently, contrastive learning frameworks \citep{hu2024comprehensive} have demonstrated that pairwise similarity objectives can significantly improve representation quality.
In particular, Supervised Contrastive Learning (SupCon) \citep{khosla2020supcon} treats all samples from the same class as positives and has been shown to outperform standard cross-entropy training on several benchmarks.

Despite these advances, existing approaches primarily rely on local pairwise relationships or class prototypes. In practice, these approaches have not performed better than cross-entropy for image classification, as cross-entropy remains the state of the art \citep{xie2020self,kolesnikov2019large}.
These approaches do not directly optimize global measures of cluster quality that simultaneously consider both intra-cluster cohesion and inter-cluster separation.
The classical clustering literature offers well-established metrics for evaluating cluster structure, which have been used for several tasks \citep{grabusts2011choice}.
Among them, the silhouette coefficient provides a simple yet powerful measure of how well each sample fits within its assigned cluster relative to neighboring clusters.
However, despite its widespread use for evaluating clustering algorithms, silhouette-based objectives have rarely been incorporated directly into representation learning.

In this work, we revisit the silhouette coefficient as a training signal for supervised representation learning.
We introduce a differentiable silhouette-based objective that can be integrated into standard deep learning pipelines and optimized jointly with classification or contrastive losses.
Intuitively, the silhouette objective encourages embeddings where samples remain close to their class neighbors while maintaining large margins from other classes.
This formulation directly promotes globally well-separated class clusters in the embedding space.

We further show that silhouette optimization complements supervised contrastive learning.
While contrastive objectives enforce pairwise similarities within a batch, the silhouette term provides a global structural signal that assesses how well each sample is positioned relative to its own class and competing classes.
Combining these two objectives yields a representation that benefits simultaneously from local pairwise alignment and global cluster-quality optimization.

In our experiments, we evaluate the proposed approach across multiple image classification datasets.
Our method consistently improves classification accuracy compared to cross-entropy, proxy-based metric learning, and supervised contrastive baselines.
When combined with multi-view supervised contrastive training, the silhouette objective yields further gains while remaining computationally efficient.

In summary, this work makes the following contributions:

\begin{itemize}
\item We introduce a differentiable silhouette-based objective for supervised representation learning that directly optimizes cluster quality in the embedding space.
\item We demonstrate that silhouette optimization is complementary to supervised contrastive learning, providing a global structural signal that improves representation geometry.
\item We empirically show consistent performance improvements across several image classification benchmarks compared to strong baselines, including cross-entropy, Proxy-NCA, Center Loss, and supervised contrastive learning.
\end{itemize}

The structure of this paper is as follows. Section \ref{sec:rel_works} reviews related works. Section \ref{sec:approach} presents a detailed description of the proposed approach. Section \ref{sec:experiments} details our experiments, encompassing datasets, methodology, and results. Lastly, Section \ref{sec:conclusion} summarizes the conclusions.

\section{Related Work}
\label{sec:rel_works}

Most deep classification models are trained using the standard cross-entropy loss, which optimizes predictive likelihood but does not explicitly enforce structure in the embedding space \citep{zhang2018generalized}.
Understanding the geometric structure of learned representations is important because these geometric characteristics directly influence generalization, robustness, transfer learning performance, and out-of-distribution detection capabilities \citep{kornblith2019similarity}.
 Models that learn well-structured geometries tend to produce embeddings in which semantically similar samples form compact regions, while distinct categories remain well separated, facilitating downstream decision boundaries and improving interpretability.
To address this limitation, several approaches introduce auxiliary losses that impose geometric constraints on learned representations \citep{wang2020understanding}.

Deep metric learning aims to learn embeddings where semantically similar samples are close and dissimilar samples are far apart.
Early approaches relied on pairwise or triplet objectives, such as contrastive loss and triplet loss.
Neighborhood-based objectives such as Neighborhood Components Analysis (NCA) \citep{goldberger2005nca} further formalize this idea by maximizing the probability that neighbors belong to the same class.
However, these approaches typically require careful sampling strategies and can scale poorly with large datasets.

Proxy-based metric learning methods address this limitation by representing each class with a learned proxy vector.
Proxy-NCA~\citep{movshovitz2017proxynca} replaces sample-to-sample comparisons with sample-to-proxy comparisons, significantly improving computational efficiency while maintaining strong retrieval performance.

Several works augment cross-entropy with auxiliary objectives that explicitly encourage compact class clusters.
Center Loss~\citep{wen2016centerloss} introduces a learnable center for each class and penalizes the distance between features and their corresponding class center.
Angular-margin losses such as ArcFace~\citep{deng2019arcface} enforce discriminative embeddings by introducing margins in angular space, leading to strong performance in face recognition and other classification tasks.
These methods improve intra-class compactness and inter-class separation but typically rely on class prototypes rather than on the global structure of the sample set.

Contrastive learning has recently emerged as a powerful framework for representation learning.
SimCLR~\citep{chen2020simclr} demonstrated that contrastive objectives combined with data augmentation can learn high-quality representations without labels.
Building on this idea, Supervised Contrastive Learning (SupCon)~\citep{khosla2020supcon} incorporates label information to treat all samples from the same class as positives.
SupCon improves representation quality and classification accuracy compared to cross-entropy on several benchmarks.
However, contrastive methods require pairwise comparisons within large batches and often rely on multiple augmented views of each sample, increasing computational cost.

Beyond pairwise or prototype-based losses, clustering metrics provide a natural way to measure representation quality.
The silhouette coefficient~\citep{rousseeuw1987silhouettes} is a classical clustering metric that quantifies how well each sample lies within its assigned cluster relative to other clusters. It measures how similar a sample is to its own cluster compared to other clusters. For a given sample $i$, the silhouette score is defined as
\begin{equation}
s(i) = \frac{b(i) - a(i)}{\max\{a(i), b(i)\}},
\end{equation}
where $a(i)$ denotes the average distance between $i$ and all other samples within the same cluster (intra-cluster distance), and $b(i)$ represents the minimum average distance between $i$ and samples from different clusters (nearest inter-cluster distance). The score ranges from $-1$ to $1$, with higher values indicating well-separated, compact clusters.

The strength of the silhouette score lies in its simultaneous assessment of cluster cohesion and separation, making it particularly effective for evaluating representation quality in embedding spaces. Unlike metrics that consider intra-class compactness and inter-class separation independently, the silhouette coefficient balances both in a single interpretable measure. This dual characterization enables robust comparisons across models and datasets and has been successfully applied to clustering validation, feature learning, and deep representation analysis \cite{arbelaitz2013extensive}. Despite their widespread use in clustering settings, silhouette-based objectives have rarely been used directly to assist supervised learning training. 

We introduce a differentiable silhouette-based objective that directly optimizes the separation between class clusters in the embedding space, specifically for supervised training. Also, our work connects the clustering-quality metric with supervised contrastive learning in a novel way.
When combined with supervised contrastive learning, the resulting objective simultaneously benefits from both pairwise contrastive structure and global cluster-quality optimization.
Empirically, we show that this combination consistently improves performance across multiple datasets while remaining computationally efficient compared to standard multi-view contrastive training.

\section{Proposed Approach}
\label{sec:approach}

We propose to improve supervised representation learning by explicitly optimizing the global structure of the embedding space.
Our method introduces a differentiable approximation of the silhouette coefficient that can be optimized jointly with supervised contrastive learning.
The resulting objective encourages representations that simultaneously satisfy local pairwise similarity constraints and global cluster separation.

\subsection{Problem Setup}

Let $\mathcal{D} = \{(x_i, y_i)\}_{i=1}^{N}$ denote a labeled dataset where $x_i \in \mathcal{X}$ is an input sample and $y_i \in \{1,\dots,C\}$ is its class label.
A neural network encoder $f_{\theta}$ maps inputs to a feature embedding:

\begin{equation}
z_i = f_{\theta}(x_i) \in \mathbb{R}^{d}.
\end{equation}

Following common practice in contrastive learning, embeddings are $\ell_2$ normalized:

\begin{equation}
\tilde{z}_i = \frac{z_i}{\|z_i\|_2}.
\end{equation}

During training, we consider mini-batches $\mathcal{B}$ containing $B$ samples.
In multi-view contrastive learning, two stochastic augmentations are applied to each input, producing a set of $2B$ embeddings.
Our goal is to learn an embedding space where samples from the same class form compact clusters while maintaining strong separation from other classes. We adopt the supervised contrastive learning objective.
Given a batch of embeddings $\{\tilde{z}_i\}$, the supervised contrastive loss for an anchor $i$ is defined as:

\begin{equation}
\mathcal{L}_{\text{sup}}^{(i)} =
-\frac{1}{|P(i)|}
\sum_{p \in P(i)}
\log
\frac{\exp(\tilde{z}_i^\top \tilde{z}_p / \tau)}
{\sum_{a \in A(i)} \exp(\tilde{z}_i^\top \tilde{z}_a / \tau)}
\end{equation}

where:

\begin{itemize}
\item $P(i)$ is the set of indices of samples sharing the same class as $i$ (positives),
\item $A(i)$ is the set of all other samples in the batch except $i$,
\item $\tau$ is a temperature parameter.
\end{itemize}

The full supervised contrastive loss over the batch is:

\begin{equation}
\mathcal{L}_{\text{sup}} =
\frac{1}{|\mathcal{B}|}
\sum_{i \in \mathcal{B}}
\mathcal{L}_{\text{sup}}^{(i)}.
\end{equation}

This objective encourages embeddings from the same class to be close while pushing apart embeddings from different classes.

\subsection{Silhouette-Based Loss}

While the contrastive objective enforces pairwise relationships, it does not directly optimize global cluster quality.
To address this limitation, we introduce a differentiable objective inspired by the silhouette coefficient.

For a sample $i$, the classical silhouette score is defined as:

\begin{equation}
s(i) = \frac{b(i) - a(i)}{\max(a(i), b(i))}
\end{equation}

where:

\begin{itemize}
\item $a(i)$ is the average distance between $i$ and samples in the same class,
\item $b(i)$ is the minimum average distance between $i$ and samples from any other class.
\end{itemize}

In our setting, we compute distances using cosine similarity:

\begin{equation}
d(i,j) = 1 - \tilde{z}_i^\top \tilde{z}_j.
\end{equation}

The intra-class distance is estimated within the batch as:

\begin{equation}
a(i) =
\frac{1}{|S(i)|}
\sum_{j \in S(i)}
d(i,j)
\end{equation}

where $S(i)$ is the set of samples in the batch sharing label $y_i$ (excluding $i$).

For inter-class separation, we compute the average distance to each class $c \neq y_i$:

\begin{equation}
d_{i,c} =
\frac{1}{|S_c|}
\sum_{j \in S_c}
d(i,j)
\end{equation}

where $S_c$ is the set of batch samples belonging to class $c$.

To approximate the minimum operator in a differentiable way, we use a soft-min formulation:

\begin{equation}
b(i) =
-\tau_s
\log
\sum_{c \neq y_i}
\exp\left(
-\frac{d_{i,c}}{\tau_s}
\right)
\end{equation}

where $\tau_s > 0$ controls the smoothness of the approximation.

\paragraph{Differentiable silhouette formulation.}
The classical silhouette definition involves the non-differentiable operator $\max(a(i), b(i))$.
To enable gradient-based optimization, we replace it with a smooth approximation based on the log-sum-exp function:

\begin{equation}
\tilde{m}(a(i), b(i)) =
\tau_m \log \left(
\exp\left(\frac{a(i)}{\tau_m}\right)
+
\exp\left(\frac{b(i)}{\tau_m}\right)
\right)
\end{equation}

where $\tau_m > 0$ is a temperature parameter.

This approximation is differentiable and converges to the exact maximum as $\tau_m \to 0$:

\begin{equation}
\lim_{\tau_m \to 0} \tilde{m}(a(i), b(i)) = \max(a(i), b(i)).
\end{equation}

Using this formulation, the differentiable silhouette score is defined as:

\begin{equation}
\tilde{s}(i) =
\frac{b(i) - a(i)}
{\tilde{m}(a(i), b(i)) + \epsilon}
\end{equation}

where $\epsilon$ is a small constant for numerical stability.

The silhouette-based loss is then given by:

\begin{equation}
\mathcal{L}_{\text{sil}} =
-\frac{1}{|\mathcal{B}|}
\sum_{i \in \mathcal{B}} \tilde{s}(i).
\end{equation}

\paragraph{Interpretation.}
This objective can be interpreted as a differentiable relaxation of the classical silhouette coefficient, encouraging representations in which intra-class distances are minimized while inter-class distances are maximized.
Unlike contrastive losses that operate at the level of individual pairs, this formulation captures a more global notion of cluster quality within each mini-batch.

\paragraph{Mini-batch estimation.}
Since the silhouette coefficient is inherently defined over the full dataset, we approximate it using mini-batches during training.
This results in a stochastic estimate of the global objective, introducing variance that depends on batch composition.

In particular, degenerate cases may arise when a batch contains very few samples from a given class, leading to unstable estimates of $a(i)$.
To mitigate this issue, we ensure that mini-batches contain multiple samples per class and analyze the impact of batch size in our experiments.

\subsection{Joint Optimization of Local and Global Structure}

Our final objective combines supervised contrastive learning with the proposed silhouette-based regularization, enabling the joint optimization of local pairwise relationships and global cluster structure:

\begin{equation}
\mathcal{L} =
\mathcal{L}_{\text{sup}}
+
\lambda_{\text{sil}}
\mathcal{L}_{\text{sil}}.
\end{equation}

The coefficient $\lambda_{\text{sil}}$ controls the contribution of the silhouette term.

\paragraph{Complementary roles of the objectives.}
The supervised contrastive loss $\mathcal{L}_{\text{sup}}$ enforces local consistency by bringing samples of the same class closer while pushing apart samples from different classes.
However, it operates at the level of individual pairs and does not explicitly account for the global arrangement of clusters.

In contrast, the silhouette-based objective $\mathcal{L}_{\text{sil}}$ evaluates each sample with respect to the structure of all classes in the batch, considering both intra-class compactness and inter-class separation.
This provides a global structural signal that complements the local nature of contrastive learning.

\paragraph{Interpretation.}
The combined objective can be interpreted as optimizing a representation space that is locally consistent and globally well-structured.
While contrastive learning shapes the local geometry of the embedding space, the silhouette term promotes coherent cluster formation and discourages overlapping class distributions.

\paragraph{Computational considerations.}
Both objectives rely on pairwise similarities within a mini-batch.
Given a batch of $M$ embeddings, the computation of the similarity matrix requires $\mathcal{O}(M^2)$ operations.
These operations can be efficiently implemented using matrix multiplication on modern accelerators.

Although the silhouette term involves class-level aggregation, it reuses the same pairwise similarity matrix computed for the contrastive loss.
As a result, it introduces only a marginal computational overhead.

\paragraph{Practical effect.}
In practice, the addition of the silhouette term stabilizes the geometry of the embedding space, leading to more compact intra-class clusters and improved separation between classes.
This effect is particularly relevant in scenarios where pairwise objectives alone may lead to fragmented or poorly separated clusters.

\section{Experiments}
\label{sec:experiments}

We evaluate the proposed silhouette-based representation learning objective across several image classification benchmarks.
Our experiments aim to answer the following questions:

\begin{itemize}
\item Does silhouette-based optimization improve classification accuracy compared to standard objectives?
\item Is the silhouette objective complementary to supervised contrastive learning?
\item Does the method consistently improve representation quality across datasets with different characteristics?
\end{itemize}

\subsection{Datasets}

Since our goal is to evaluate if our method consistently improves representation quality across datasets with different characteristics, we performed the experiments on seven widely used datasets for classification benchmark: CIFAR-10\citep{krizhevsky:09}, CIFAR-100 \citep{krizhevsky:09}, Stanford Cars \citep{KrauseStarkDengFei:13}, Caltech-101 \citep{FeiFei2004LearningGV}, Caltech-256 \citep{griffin2007caltech}, FGVC-Aircraft \citep{maji13fine-grained}, and Oxford Flowers \citep{Nilsback08}.
All datasets contain color images representing diverse contexts, emphasizing different categories and exhibiting distinct characteristics. While some datasets involve fine-grained classification tasks, others differentiate categories at varying levels of granularity. This diversity enables a more comprehensive evaluation of the proposed approach.

\subsection{Methodology}

We evaluate our loss in the supervised image classification task by measuring the top-1 and top-5 accuracy in the previously presented datasets. Our approach yielded two training strategies: Cross-entropy with silhouette regularization (CE+Sil) and Cross-entropy with SupCon2 combined with silhouette regularization (CE+SupCon2+Sil). We compare our approach with the following state-of-the-art training strategies: classical Cross-entropy (CE) \citep{mao2023cross}, Supervised contrastive learning (SupCon) and Two-view supervised contrastive learning (SupCon2) \citep{khosla2020supcon}, Proxy-NCA \citep{movshovitz2017proxynca}, and Center Loss (Center) \citep{wen2016centerloss}.

All experiments use a convolutional encoder (EfficientNet B0 \citep{tan2019efficientnet}) followed by a projection head used during representation learning. The embeddings are $\ell_2$ normalized before computing contrastive and silhouette objectives. We used the AutoAugment \citep{cubuk2019autoaugment} strategy, a learning rate of 0.001, weight decay of 0.0001, batch size of 256, dropout of 0.2, and trained for 100 epochs. Final performance is reported on the test set using the checkpoint of the last epoch.

\subsection{Results}

\begin{figure}
\centering
\includegraphics[width=.9\textwidth]{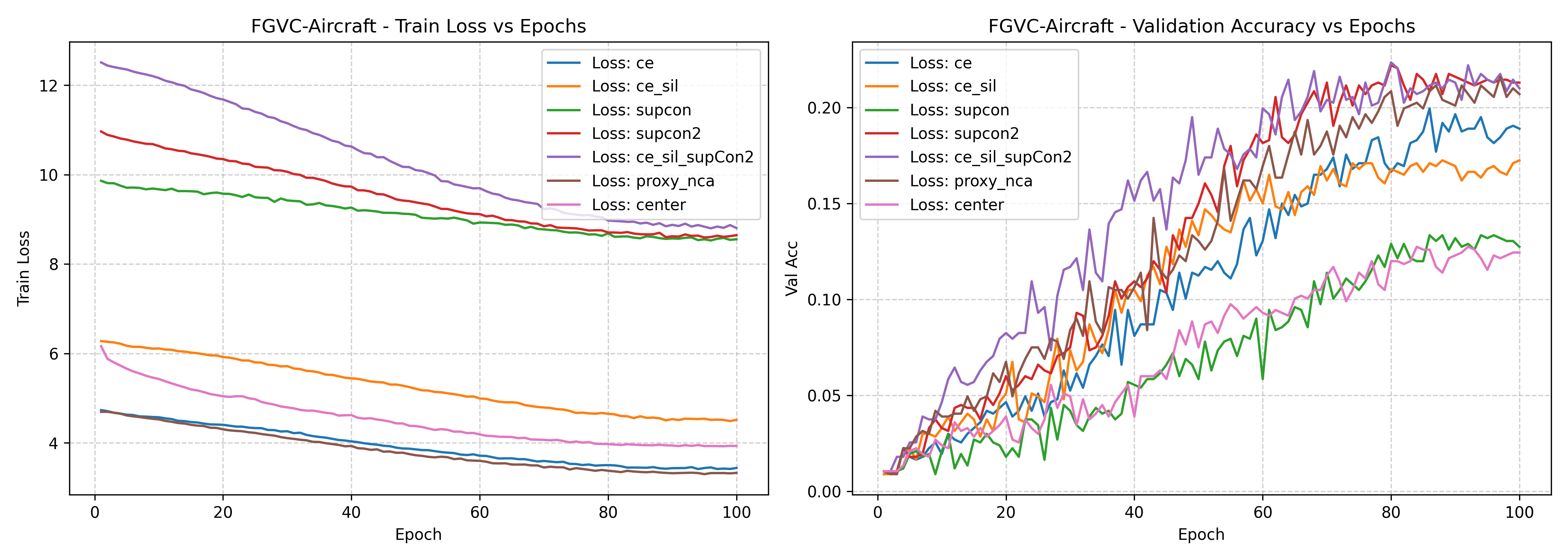}
\caption{Training loss and validation accuracy curves across training epochs for the FGVC-Aircraft dataset.}
\label{fig:aircraft}
\end{figure}

\begin{figure}
\centering
\includegraphics[width=.9\textwidth]{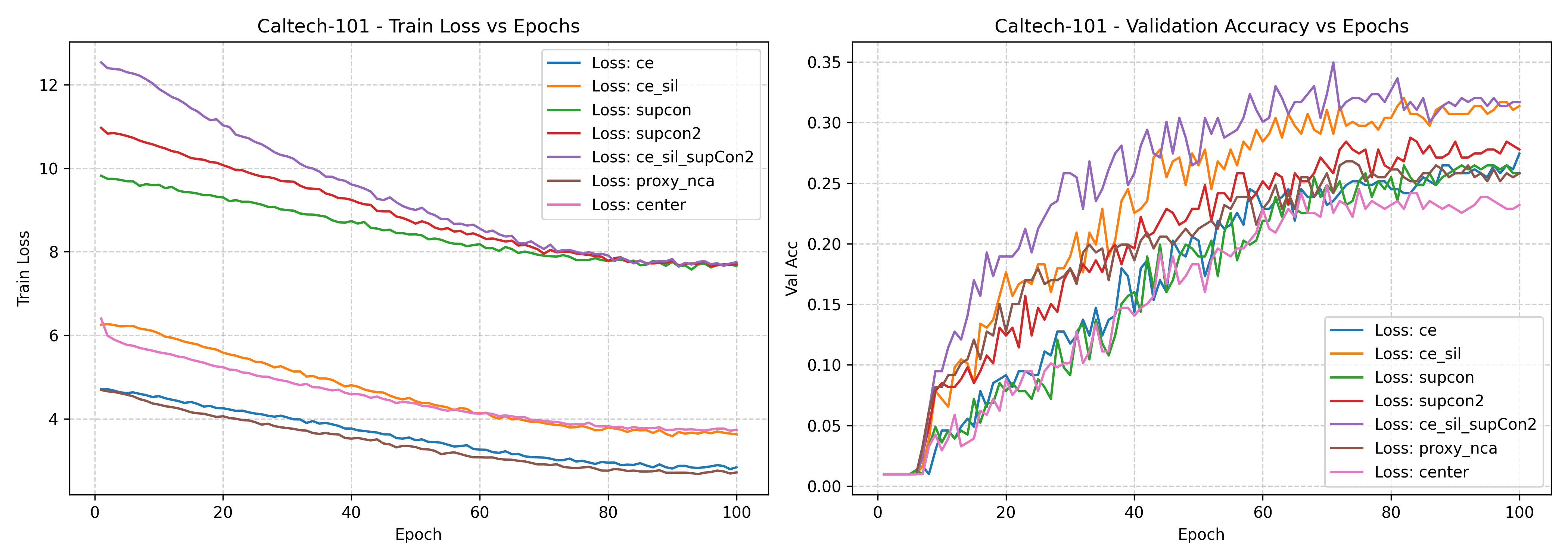}
\caption{Training loss and validation accuracy curves across training epochs for the Caltech-101 dataset.}
\label{fig:caltech}
\end{figure}

\begin{figure}
\centering
\includegraphics[width=.9\textwidth]{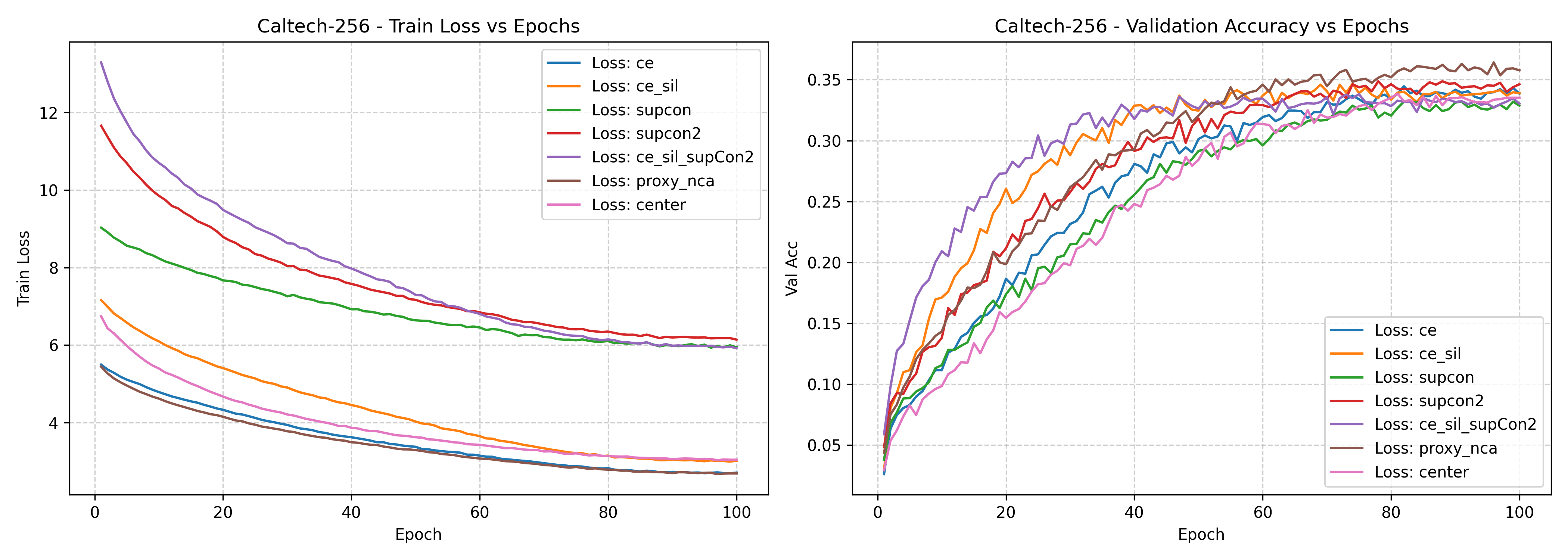}
\caption{Training loss and validation accuracy curves across training epochs for the Caltech-256 dataset.}
\label{fig:caltech256}
\end{figure}

\begin{figure}
\centering
\includegraphics[width=.9\textwidth]{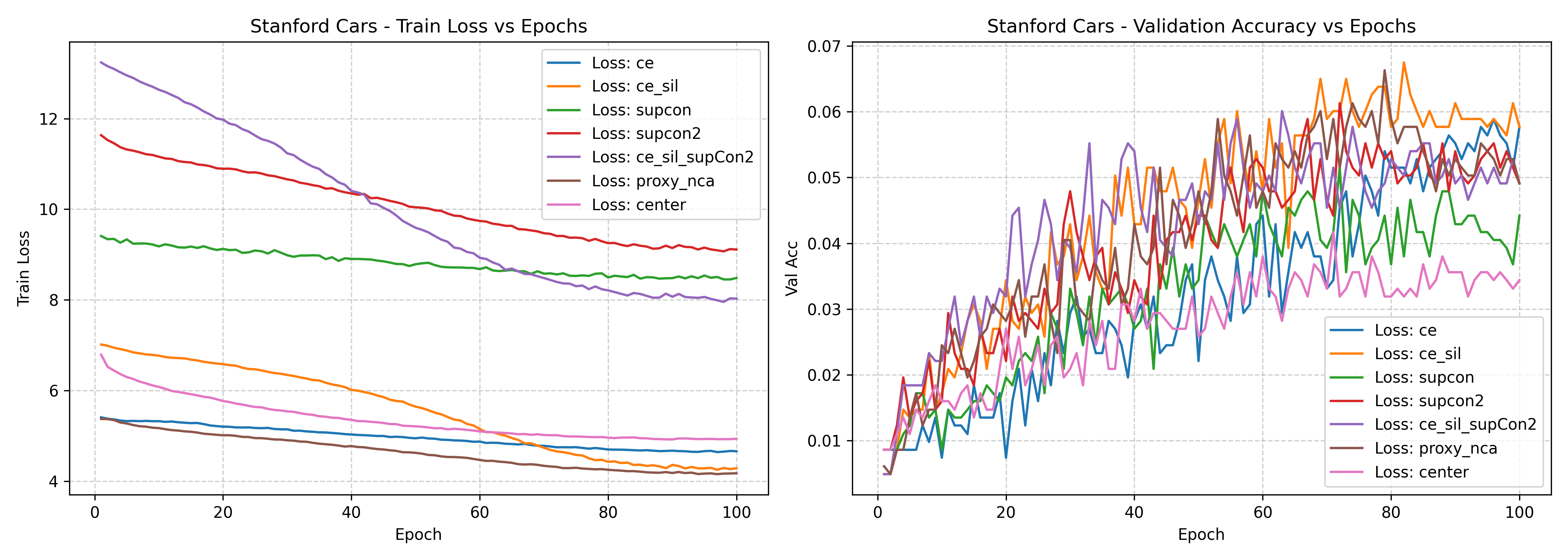}
\caption{Training loss and validation accuracy curves across training epochs for the Stanford Cars dataset}
\label{fig:cars}
\end{figure}

\begin{figure}
\centering
\includegraphics[width=.9\textwidth]{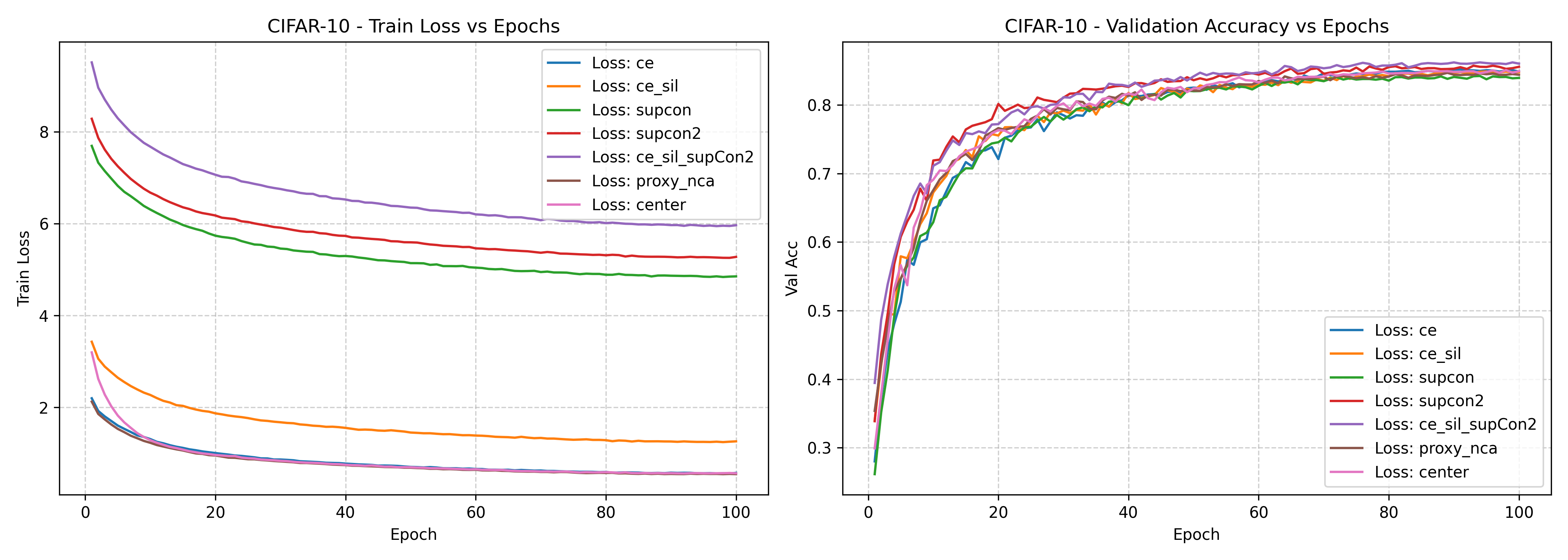}
\caption{Training loss and validation accuracy curves across training epochs for the CIFAR-10 dataset.}
\label{fig:cifar10}
\end{figure}

\begin{figure}
\centering
\includegraphics[width=.9\textwidth]{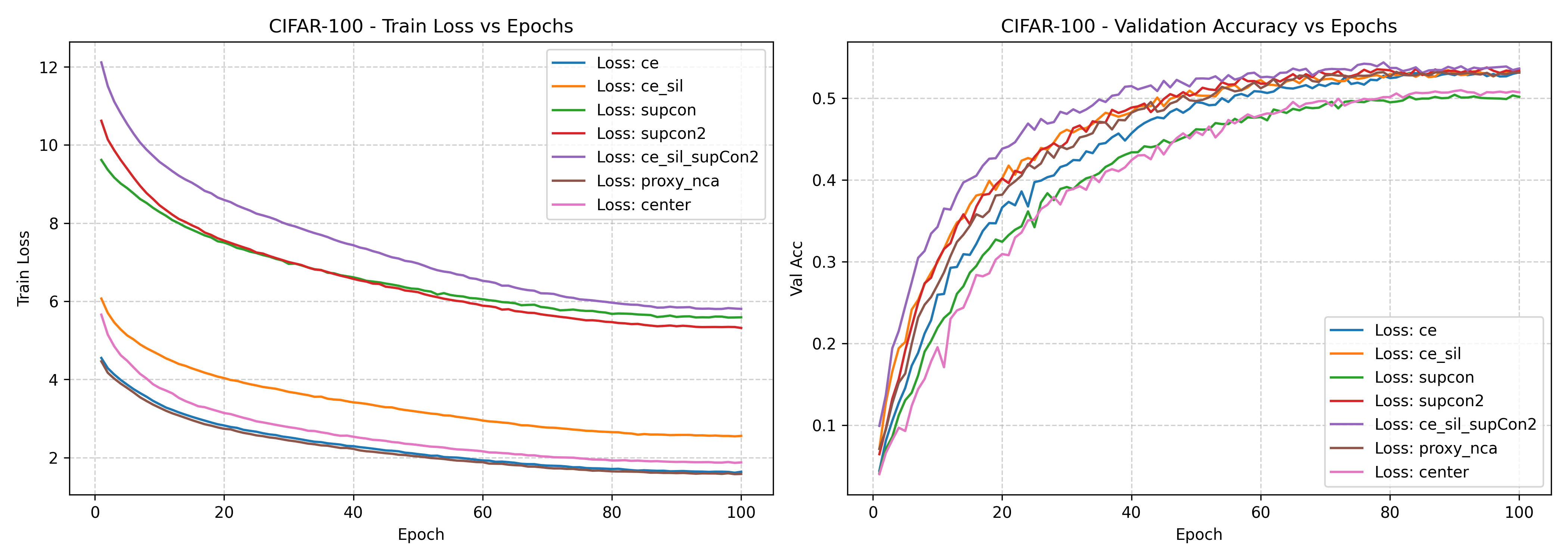}
\caption{Training loss and validation accuracy curves across training epochs for the CIFAR-100 dataset.}
\label{fig:cifar100}
\end{figure}

\begin{figure}
\centering
\includegraphics[width=.9\textwidth]{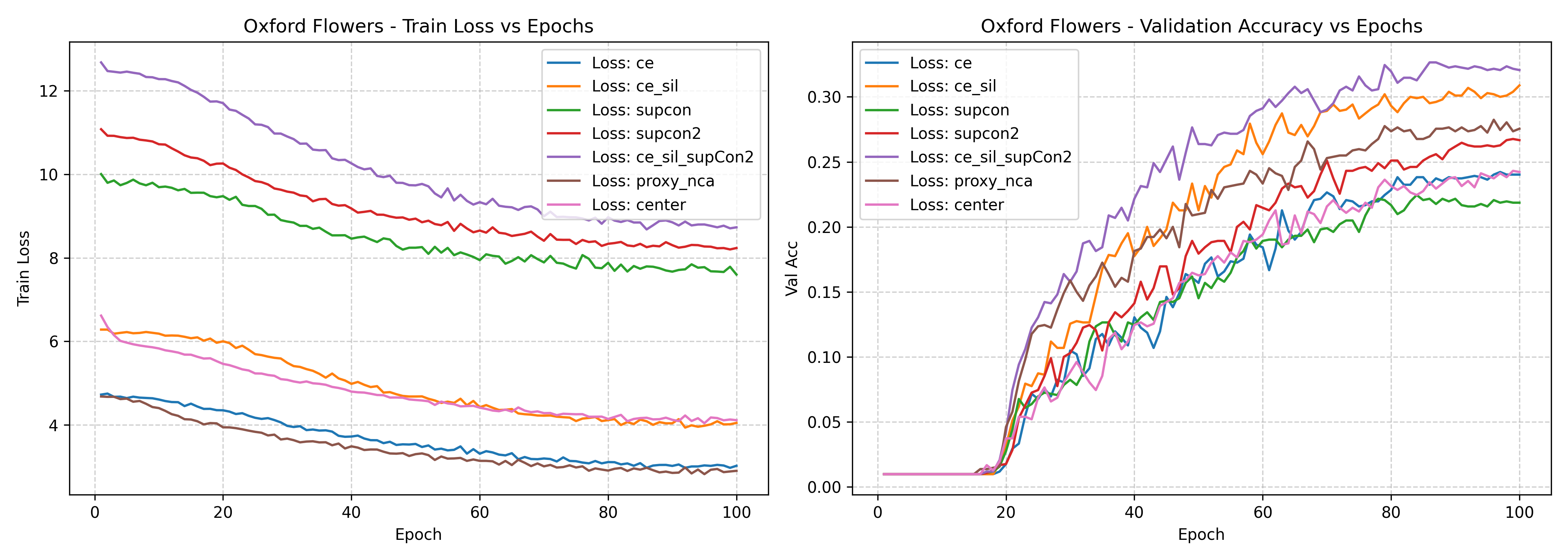}
\caption{Training loss and validation accuracy curves across training epochs for the Oxford Flowers dataset.}
\label{fig:flowers}
\end{figure}

Table \ref{tab:results} presents the Top-1 and Top-5 of each loss across all evaluated datasets.
Figures~\ref{fig:aircraft}--\ref{fig:flowers} summarize the training loss and the validation accuracy per epoch of each loss across each dataset. The proposed combination of supervised contrastive learning and silhouette regularization achieves the best performance, improving Top-1 test accuracy by +4.11\% over cross-entropy, +3.10\% over CE+Sil, +2.12\% over SupCon2, and +3.19\% over Proxy-NCA. These results suggest that the silhouette objective provides a complementary signal that improves representation learning beyond both prototype-based and contrastive methods.

\begin{table*}
\centering
\caption{Top-1 and top-5 classification accuracy across the datasets.}
\label{tab:results}
\resizebox{\textwidth}{!}{%
\begin{tabular}{lcccccccccccccc}
\hline
 & \multicolumn{2}{c}{CE} & \multicolumn{2}{c}{CE+SIL} & \multicolumn{2}{c}{SupCon} & \multicolumn{2}{c}{SupCon2} & \multicolumn{2}{c}{CE+SIL+SupCon2} & \multicolumn{2}{c}{ProxyNCA} & \multicolumn{2}{c}{Center} \\
Dataset & Top1 & Top5 & Top1 & Top5 & Top1 & Top5 & Top1 & Top5 & Top1 & Top5 & Top1 & Top5 & Top1 & Top5 \\
\hline

CIFAR-10
& 0.8391 & 0.9938
& 0.8393 & \textbf{0.9942}
& 0.8437 & 0.9924
& \underline{0.8505} & 0.9939
& \textbf{0.8514} & \underline{0.9941}
& 0.8446 & --
& 0.8438 & 0.9934 \\

CIFAR-100
& 0.5283 & 0.8057
& \underline{0.5322} & 0.8068
& 0.5042 & 0.7981
& 0.5378 & \textbf{0.8144}
& \textbf{0.5386} & \underline{0.8068}
& 0.5305 & --
& 0.5070 & 0.7977 \\

Caltech-101
& 0.4404 & 0.6347
& \underline{0.4595} & 0.6383
& 0.3857 & 0.6007
& 0.4417 & \underline{0.6404}
& \textbf{0.4692} & \textbf{0.6631}
& 0.4416 & --
& 0.3832 & 0.5709 \\

FGVC-Aircraft
& 0.1776 & \underline{0.4821}
& 0.1704 & 0.4341
& 0.1110 & 0.3738
& \textbf{0.2109} & \textbf{0.5209}
& \underline{0.2043} & 0.4785
& 0.1947 & --
& 0.1236 & 0.3966 \\

Oxford Flowers
& 0.1960 & 0.4721
& \underline{0.2656} & \underline{0.5456}
& 0.1791 & 0.4654
& 0.2106 & 0.5025
& \textbf{0.2740} & \textbf{0.5611}
& 0.2210 & --
& 0.2018 & 0.4789 \\

Caltech-256
& 0.3391 & \textbf{0.5521}
& 0.3278 & 0.5230
& 0.3228 & 0.5279
& \underline{0.3422} & \underline{0.5480}
& 0.3393 & 0.5278
& \textbf{0.3549} & --
& 0.3254 & 0.5328 \\

Stanford Cars
& 0.0491 & 0.1505
& \underline{0.0596} & \textbf{0.1663}
& 0.0484 & 0.1578
& 0.0556 & 0.1566
& 0.0585 & \underline{0.1591}
& \textbf{0.0648} & --
& 0.0387 & 0.1384 \\

\hline
Average
& 0.3671 & 0.5844
& \underline{0.3792} & 0.5869
& 0.3421 & 0.5594
& 0.3785 & \underline{0.5967}
& \textbf{0.3908} & \textbf{0.5980}
& 0.3789 & --
& 0.3320 & 0.5580 \\

\hline
\end{tabular}
}
\end{table*}

Several key trends emerge from these results. While adding silhouette regularization to cross-entropy (CE+SIL) improves accuracy on half of the datasets, indicating that explicit cluster optimization can sometimes benefit standard classification, the most significant gains occur when silhouette regularization is combined with two-view supervised contrastive learning (CE+SIL+SupCon2). This combined approach consistently ranks among the top-performing methods, yielding the best results across multiple datasets and the highest average top-1 and top-5 accuracies.

The lack of consistent improvement from CE+SIL alone, in contrast to the success of CE+SIL+SupCon2, suggests that supervised contrastive learning and silhouette optimization provide complementary learning signals. Specifically, supervised contrastive learning enforces local pairwise alignment within a batch, whereas the silhouette objective evaluates a sample's position relative to the global class distribution. We hypothesize that SupCon enhances local neighborhood structure, while the silhouette objective improves global cluster separation. Consequently, combining these objectives yields embeddings that are both locally coherent and globally well-structured, which explains the consistent performance gains across datasets. However, it is important to note that SupCon has a substantial computational cost compared with SIL term calculation, which justifies using CE+SIL in certain situations.

An important observation from the figures is that the CE+SIL+SupCon2 loss consistently achieves higher validation accuracy during the early training epochs than the other loss functions. As training progresses, however, the performance differences gradually diminish, and near the final epochs, the validation accuracy converges, stabilizing at levels comparable to the other losses. This is more evident in the Caltech-101 and FGVC-Aircraft datasets, shown in Figures \ref{fig:caltech} and \ref{fig:aircraft}, respectively.

Datasets such as Caltech-101 and Oxford Flowers exhibit particularly strong gains.
One has more generic classes, and the other is a fine-grained dataset, showing the potential of the loss in two different situations. 

Overall, the results demonstrate that explicitly optimizing cluster-quality metrics can significantly improve supervised representation learning.
Despite its conceptual simplicity, the silhouette objective provides a strong global structural signal that complements existing representation learning objectives.
The proposed method consistently improves performance across datasets while introducing minimal computational overhead.

\section{Conclusion}
\label{sec:conclusion}

In this work, we revisited the silhouette coefficient as a training signal for supervised representation learning.
While silhouette has long been used as a metric for evaluating clustering quality, its potential as a differentiable objective for shaping representation spaces has remained largely unexplored in deep learning. We introduced a differentiable silhouette-based loss that can be integrated into standard representation-learning pipelines and jointly optimized with existing objectives.
The proposed formulation encourages embeddings that simultaneously achieve low intra-class distances and strong inter-class separation.
When combined with supervised contrastive learning, the silhouette objective provides a complementary global structural signal that improves the geometry of the learned embedding space.

Extensive experiments across multiple image classification benchmarks demonstrate that the proposed approach consistently improves performance compared to strong baselines, including cross-entropy, proxy-based metric learning, center loss, and supervised contrastive learning.
In particular, combining the silhouette objective with two-view supervised contrastive training yields the strongest results, achieving substantial gains over both standard classification training and contrastive learning alone.
These findings suggest that explicitly optimizing cluster-quality metrics can provide meaningful benefits for supervised representation learning.
Our results further indicate that supervised contrastive learning and silhouette optimization address complementary aspects of representation quality.
While contrastive learning enforces local pairwise similarity constraints within a batch, the silhouette objective evaluates how well each sample is positioned relative to the entire class clusters.
The combination of these signals leads to embeddings that are both locally coherent and globally well-separated, resulting in improved classification performance across diverse datasets.

Despite the promising results, several directions remain for future work.
First, a deeper analysis of the interaction between contrastive learning and cluster-quality objectives could provide further insight into the geometric properties induced by the proposed loss.
In particular, ablation studies investigating the influence of the silhouette weighting coefficient, temperature parameters, and batch size would help characterize the approach's robustness.
Second, additional experiments evaluating representation geometry could further validate the proposed method.
For example, embedding visualizations using dimensionality reduction techniques such as t-SNE or UMAP may provide qualitative evidence of improved cluster structure.
Similarly, measuring additional clustering metrics beyond silhouette scores could help better understand the relationship between cluster geometry and downstream task performance.
Third, the computational properties of the proposed objective warrant further investigation.
Although the silhouette term introduces minimal overhead in our current implementation, exploring efficient approximations for larger batch sizes or very large datasets could enable broader applicability in large-scale training settings.

Finally, extending the silhouette-based objective beyond supervised classification represents an interesting avenue for future research.
Potential directions include semi-supervised representation learning, self-supervised contrastive frameworks, and metric learning scenarios where explicit cluster structure is beneficial.
Overall, our results highlight the potential of cluster-quality optimization as a complementary paradigm for representation learning.
We hope this work encourages further exploration of clustering-inspired objectives to improve the structure and utility of learned representations.

\bibliographystyle{unsrtnat}
\bibliography{references}

\end{document}